\documentclass{article}
\usepackage{graphicx} 
\usepackage{comment}
\usepackage[numbers]{natbib}
\usepackage{url}

\usepackage{booktabs}
\usepackage{authblk}

\usepackage{xcolor}

\title{Beyond Cox Models: Assessing the Performance of Machine-Learning Methods in Non-Proportional Hazards and Non-Linear Survival Analysis} 
\date{Apr 2025}

\author[1,*,\dag]{Giovanni Birolo}
\author[1]{Ivan Rossi}
\author[1]{Flavio Sartori}
\author[1]{Cesare Rollo}
\author[1,*]{Piero Fariselli}
\author[1,*]{Tiziana Sanavia}

\affil[1]{Department of Medical Sciences, University of Turin, Turin, Italy}
\affil[*]{These authors jointly supervised this work.}
\affil[$\dag$]{Corresponding author: giovanni.birolo@unito.it}

\begin{document}

\maketitle

\section{Abstract}

Survival analysis often relies on Cox models, assuming both linearity and proportional hazards (PH). This study evaluates machine and deep learning methods that relax these constraints, comparing their performance with penalized Cox models on a benchmark of three synthetic and three real datasets. In total, eight different models were tested, including six non-linear models of which four were also non-PH.
Although Cox regression often yielded satisfactory performance, we showed the conditions under which machine and deep learning models can perform better. Indeed, the performance of these methods has often been underestimated due to the improper use of Harrell's concordance index (C-index) instead of more appropriate scores such as Antolini's concordance index, which generalizes C-index in cases where the PH assumption does not hold. In addition, since occasionally high C-index models happen to be badly calibrated, combining Antolini's C-index with Brier's score is useful to assess the overall performance of a survival method. Results on our benchmark data showed that survival prediction should be approached by testing different methods to select the most appropriate one according to sample size, non-linearity and non-PH conditions. To allow an easy reproducibility of these tests on our benchmark data, code and documentation are freely available at \url{https://github.com/compbiomed-unito/survhive}.

\section{Introduction}

Survival analysis represents a collection of statistical and machine learning procedures where the outcome variable of interest is a time-to-event, representing not only whether or not an event occurred, but also when that event occurred, which is common in several contexts, ranging from component failure in engineering to adverse events in medicine. The prediction of these time-to-event cases makes survival analysis much more challenging than other analyses like regression or classification. Another important aspect of survival analysis is the presence of censored cases, i.e. when the observed event does not occur during the study’s timeframe (for example, death or disease relapse after the end of a clinical study). 

The Cox proportional hazards (CoxPH) model \cite{Cox1972} is by far the most well known and widely used statistical approach in survival analysis, which extends linear regression to survival outcomes and assumes that the hazard functions of all individuals are proportional (PH), greatly simplifying the fit of the model by turning it into a regression problem. Furthermore, CoxPH often achieves high predictive performance, showing that it is reasonably robust to minor deviations from its underlying assumptions. Moreover, being also directly interpretable, it is not surprising that it has remained the most popular model in survival analysis for the last fifty years since its introduction. 

However, datasets that are strongly non-linear and/or non-PH do exist and this motivated the development of new methods based on more recent techniques.
Thus, successful machine learning methods were adapted to survival analysis, such as CoxNet \cite{Simon2011}, a version of the CoxPH regression with ElasticNet regularization, Random Survival Forests (RSF) \cite{Ishwaran2008} and Gradient Boosting Survival Analysis (GrBSA) \cite{Hothorn2005} models, based on Random Forests and Gradient Boosting algorithms, respectively. More recently, many deep learning survival methods have been developed to address this task. The first has been DeepSurv \cite{Katzman2018}, a straightforward deep learning extension of CoxPH allowing non-linearity while preserving the PH assumption and the Cox loss. Other deep versions of the CoxPH model have been introduced after DeepSurv, for instance FastCPH \cite{Yang2022}, an extension of the LassoNet\cite{Lemhadri2021} method to survival outcomes. DeepHit \cite{Lee2018} takes a more radical departure from CoxPH, predicting a discrete probability distribution on binned event times, avoiding the PH assumption and other parametric distributions. Deep Survival Machines \cite{Nagpal2021} aim to model the underlying event outcome distribution as a mixture of some fixed parametric (Weibull or Log-normal) distributions, whose parameters, as well as the mixing weights, are modeled using deep neural networks. Another survival model based on deep learning, SurvTRACE \cite{Wang2022}, employs a transformer that incorporates an attention mechanism instead of the more traditional network architectures used by the previous methods.

Machine learning and especially deep learning survival predictors advertise themselves as being more powerful than CoxPH, claiming to provide more accurate predictions on real datasets, when linearity and PH assumptions are not guaranteed. However, they also present their own disadvantages, ranging from their substantially larger computational costs to the need for more data and the requirement for complex training strategies, often coupled with a reduced interpretability of the results.

Another critical aspect affecting survival analysis is the choice of the evaluation metrics for predictive models. The Concordance Index (C-index) is one of the most used metrics, that quantifies the rank correlation between the actual survival times and the predictions of a model, accounting for censoring. Multiple estimators have been proposed, but the popularity of the PH assumption led to the adoption of Harrell's C-index \cite{Harrell1996} as the de-facto standard metric. It measures the ability of a survival model to order subjects by estimating the proportion of correctly ordered pairs of individuals among all comparable pairs in the dataset. Individuals with higher risk scores have a higher probability of an event at every time value. This means that the ranking is fixed between individuals and is the same at all times. However, in non-PH models, the ranking of individuals is not necessarily fixed and can change depending on the time, i.e. the survival curves predicted for two individuals can cross. In these cases, a generalization of Harrell's C-index must be used to account for the rank dependence on the time. To this aim, Antolini's C-index \cite{Antolini2005} has been introduced as an appropriate evaluation metric for non-PH models. Nevertheless, Harrell's C-index has been used, incorrectly, to evaluate non-PH models, often trying to partially account for the time dependence of ranking by averaging the C-index scores computed from the survival probability at several fixed times.

Therefore, we decided to better investigate advantages and drawbacks of the most recent survival analysis methods by performing their systematic comparison on six different datasets, including three synthetic datasets designed to violate some (or all) of CoxPH's base assumptions and three real clinical datasets. We also explored how the choice of the evaluation metric affects the results. Our study provides insights into the conditions under which machine learning approaches can outperform CoxPH, highlighting the importance of the type of data in determining the most suitable survival analysis methodology. To allow an easy reproducibility of our analyses on different datasets, we organized the code used in this work into the SurvHive package \cite{birolo2025}, a Python-based framework developed to provide a unified and consistent API for executing several survival analysis methods, which often differ in input/output formats and capabilities. Compatible with both \textit{scikit-learn} \cite{pedregosa2011,buitinck2013} and \textit{scikit-survival} \cite{Poelsterl2020} libraries, it standardizes data handling, supports hyperparameter tuning, time-dependent risk evaluation, and cross-validation for censored data. By using adapters instead of re-implementing methods, SurvHive minimizes bias and preserves the integrity of original models. Code and documentation are freely available at \url{https://github.com/compbiomed-unito/survhive}.

\section{Methods}
\subsection{Survival methods}

We tested eight different survival methods. Four of them are from the \textit{scikit-survival} library: the classical CoxPH \cite{Tibshirani1997} and the machine learning-based CoxNet \cite{Simon2011} (which adds the ElasticNet regularization to CoxPH), GrBSA \cite{Hothorn2005} (GradientBoostingSurvivalAnalysis using the CoxPH loss and thus also assuming PH) and RSF \cite{Ishwaran2008} (RandomSurvivalForest, a non-PH method).

The other four deep learning methods have different sources: 
\begin{itemize}
    \item DeepHit \cite{Lee2018}, implemented by the DeepHitSingle class from \textit{PyCox} \cite{Kvamme2019} package;
    \item Deep Survival Machines (DSM) \cite{Nagpal2021}, from \textit{Auton Survival} \cite{Nagpal2022} package;
    \item FastCPH \cite{Yang2022}, from \textit{LassoNet} \cite{Lemhadri2021} package;
    \item SurvTRACE, implemented by the SurvTRACESingle class from \textit{SurvTRACE} \cite{Wang2022} package.
\end{itemize}

\subsection{Datasets}
In order to compare the methods, we used six different datasets. 
Three of them are synthetic datasets specifically designed to test the limits of the Cox-based methods, while the others are real data coming from clinical studies.

\subsubsection{Synthetic datasets}

We generated the three synthetic datasets following different assumptions:

\begin{description}
    \item[LinPH]: proportional hazards from a linear combination of features,
    \item[NonLinPH]: proportional hazards from a non-linear function of the features,
    \item[NonPH]: non proportional hazards.
\end{description}
The second and third datasets violate one or both CoxPH assumptions and are meant to stress-test the methods, in order to better characterize their limitations.

All three datasets have 20 features \(x_1, \ldots, x_{20}\), whose value is randomly sampled from a normal distribution. 
Events are randomly sampled from a \([0, 10]\) time interval according to the survival function estimated from the features. 

Both LinPH and NonLinPH generative models have a baseline hazard that is generated from random noise in a bounded time interval. Specifically, LinPH model computes the risk from a linear combination of the first 8 features with random coefficients, while NonLinPH model uses a fixed function of the first 4 features: \(x_1 + x_2 x_3 + \cos(6 x_4)\). Finally, NonPH model splits the time interval into 16 same-sized intervals and yields survival functions that are piecewise constant on the intervals. The probability $p_i$ of an event for each interval is computed using the soft-max function on the first 16 features (one for each time interval) multiplied by 16: 
$p_i = \frac{\exp(16 x_i)}{\sum^{16}_{j=1} \exp(16 x_j)}$.

The synthetic datasets were generated with the goal of strongly violating the linearity and PH assumptions while also being highly predictable, meaning that the event time is mostly determined by the features and not overly affected by the random sampling.
In this way, a good model will be able to achieve high predictive performance, allowing us to discriminate better between the different models.

\subsubsection{Clinical datasets}

In addition to the synthetic datasets described above, we also considered three publicly available survival datasets, commonly used to assess the performance of predictive models: 
\begin{description}
    \item[FLChain]: the Assay of Serum Free Light Chain \cite{Therneau2015}, including 6,524 samples and 8 features;
    \item[GBSG2]: the Rotterdam tumour bank and German Breast Cancer Study Group \cite{Foekens2000,Schumacher1994}, including 3,668 samples and 8 features; 
    \item[METABRIC]: the Molecular Taxonomy of Breast Cancer International Consortium \cite{Curtis2012}, including 1,903 samples and 9 features.
\end{description}


FLChain and METABRIC datasets were preprocessed according to Kvamme et al. \cite{Kvamme2019}, while for GBSG2 the same preprocessing schema of Royston and Altman \cite{Royston2013} was applied, with the only difference that the data from the two studies were merged to create a single dataset. 

\subsection{Metrics}

Model performance was evaluated using all the most commonly used metrics in survival analysis:
\begin{enumerate}
    \item Harrell's C-index \cite{Harrell1996}, calculated at three time quartiles (25\%, 50\%, 75\%);
    \item Antolini's Concordance Index \cite{Antolini2005}, the extension of Harrell's C-index for models with time-dependent risk ranking;
    \item Brier score \cite{Brier1950}, calculated at multiple time quantiles;
    \item Area under the receiver operator curve (AUROC) at multiple time quantiles.
\end{enumerate}
Note that Harrell's C-index is not appropriate for models where the risk ranking between individuals is not fixed (that is, when survival curves can cross each other). All PH models have fixed ranking, but for non-PH models this is generally not the case.
However, we computed it for all models to show the impact of using Harrell's C-index instead of Antolini's when the comparison includes non-PH models.

\subsection{Testing}
We used a nested cross-validation approach for the hyper-parameter optimization and the model evaluation.
For each dataset, model performance was estimated using an outer 3-fold cross-validation. For each fold, independent hyper-parameter optimization was performed on the training set with a full grid search using an inner 5-fold cross-validation, which was repeated twice with reshuffling, for a total of 10 validation sets. The hyper-parameters' combination yielding the highest average Antolini's C-index was then selected to refit the model on the whole training set. Finally, the fitted model was evaluated on a hold-out set for each of the three test sets of the outer cross-validation.

All computations were performed on a virtual machine with 12 Intel Xeon Skylake processors and 1 Nvidia Tesla T4 GPU. The parameter-optimization runs were performed using 12 concurrent processes for the non-GPU methods and a single process for the GPU-accelerated ones.

\section{Results}
Table \ref{tab:methods} reports the tested survival methods, highlighting whether they follow one or both fundamental CoxPH assumptions, i.e. the hazard functions of all individuals are proportional (PH assumption) and the proportionality is computed through a linear combination of the features (linearity assumption). 

\begin{table}[ht]
 \centering
    \begin{tabular}{lcc}
        \toprule
        \textbf{Method} & \textbf{Linearity} & \textbf{PH}\\
        \midrule
        CoxPH/CoxNet            & Yes       & Yes\\
        GrBSA               & No        & Yes\\
        RSF                     & No        & No\\
        FastCPH                 & No        & Yes\\
        DeepHit                 & No        & No\\
        DeepSurvivalMachines (DSM)    & No        & No\\
        SurvTRACE               & No        & No\\
        \bottomrule
    \end{tabular}
    \caption{\textbf{Tested methods and their characterization.}}
    \label{tab:methods}
\end{table}


The three synthetic datasets satisfy either both assumptions (LinPH), or only PH (NonLinPH), or none of them (NonPH), providing a ``controlled'' environment to evaluate the methods. On the other hand, checking the validity of these assumptions on the real datasets is not straightforward, as we do not directly observe the complete survival distribution of each patient but only the final event time. Therefore, we used the synthetic datasets to guide the interpretation of the results observed in the clinical datasets FLChain, GBSG2, and Metabric. 

\subsection{Concordance index}

\begin{figure}[t]
    \centering
    \includegraphics[width=1.0\linewidth]{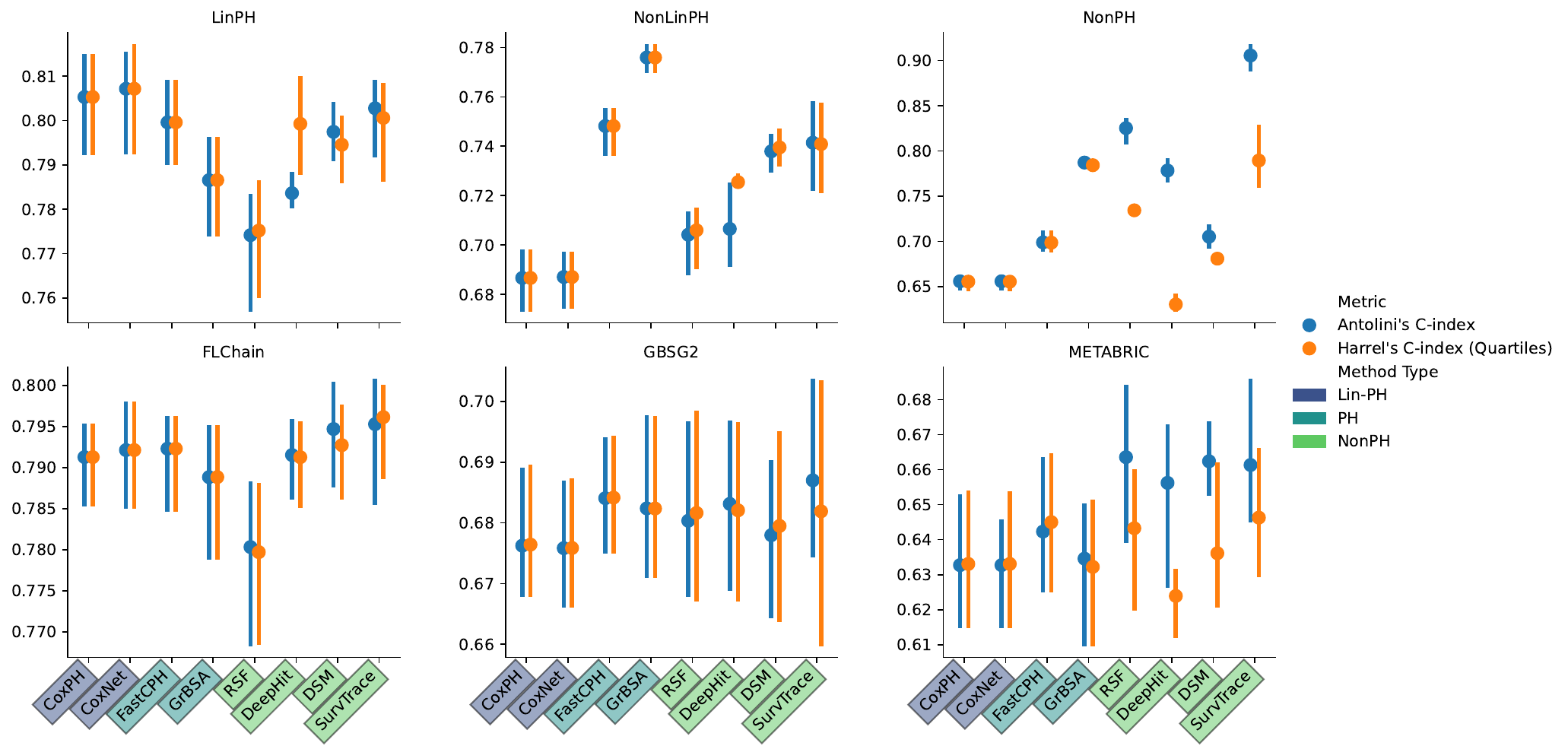}
    \caption{\textbf{C-index performance across all methods and datasets.} Error bars are displayed for both Harrell's (orange bars) and Antolini's (blue bars) C-indices. Specifically, each dot represents the average C-index, while the bars show the C-index range resulting from the 3-fold cross-validation. Methods' assumptions are highlighted with different colors in the horizontal labels.}
    \label{fig:antolini-harrel-all}
\end{figure}

For each dataset, the models' performance has been evaluated three times with a 3-fold cross-validation. As discussed above, Harrell's C-index was developed in a PH context, assuming a fixed risk ranking between individuals. Unfortunately, it has often been incorrectly used for evaluating non-PH models, often taking its average at multiple times in an attempt to account for the  rank dependence upon time. However, this is not enough: for comparisons including non-PH models, Antolini's C-index should be used, since it generalizes Harrel's C-index to models without fixed risk ranking. To show the bias introduced by this misuse, we computed both Antolini's and Harrell's C-indices (with Harrel's computed as an average on the first, second and thirds quartiles of event times). The results for all datasets and methods are plotted in Figure \ref{fig:antolini-harrel-all} and full data is reported in Supp. Table 1. 

On the synthetic LinPH dataset, CoxPH and CoxNet methods are the top performers as expected. In principle, other more powerful models should be able to reach similar scores, but only FastCPH and SurvTRACE show close performance to the Cox-based methods. Both Harrell's and Antolini's C-indices are quite similar for all methods in this dataset, except for DeepHit, whose predictions seem to deviate significantly from PH.

In the NonLinPH dataset, non-linear methods achieved much higher performance than the linear CoxPH and CoxNet. Here, the best performing methods are FastCPH and GrBSA, which are PH methods with non-linear risks, with GrBSA showing the highest performance. FastCPH performed only slightly better than DSM and SurvTRACE. DeepHit and RSF do better than the linear models too.

The gap in performance among the methods becomes very large on the synthetic NonPH dataset. Both CoxPH and CoxNet achieved an average C-index of 0.66, with no significant differences between Harrell's and Antolini's metrics. Despite the assumptions of these two methods are strongly violated in this dataset, they are still able to provide some prediction power. On the other hand, SurvTRACE, which does not make any assumption, was able to achieve an Antolini's C-index of 0.9. It is also worth highlighting that GrBSA, which assumes PH, almost reached an Antolini's C-index of 0.8, showing that some PH methods can still perform well even when PH does not hold.
In this dataset, we were also able to observe large differences between the C-indices for the non-PH methods, showing a Harrell's C-index much lower than Antolini's metric. According to our expectations, Harrell's C-index strongly penalizes non-PH methods, making the performance of SurvTRACE seem comparable to those of GrBSA, while Antolini's shows a 0.1 of advantage for SurvTRACE.

Turning our attention to the real clinical datasets, almost all tested methods achieved high C-indices on the FLChain dataset, without significant differences between Harrell's and Antolini's metrics, suggesting a behavior similar to that in the LinPH synthetic dataset, except for RSF, which performed visibly worse, albeit by a small margin. Even if both SurvTRACE and DSM appear to be on average slightly better than the others, the good performance of both CoxPH and CoxNet suggests that this dataset does not heavily deviate from linearity and PH assumptions. 

Survival methods applied to GBSG2 clinical dataset performed almost all equally well, with a slight advantage for non-linear methods. Low differences between Harrell's and Antolini's C-index observed in non PH methods, i.e. FastCPH and GrBSA, suggest that the PH assumption does not seem restrictive in this dataset.

Conversely, in METABRIC the non-PH methods are the best performing, even though this advantage is appreciable only when using Antolini's C-index, while Harrell's C-index values are comparable to those observed for the other methods. PH methods performed noticeably worse, with a slight advantage for non-linear PH methods (FastCPH). We can conclude that PH is a limiting assumption in METABRIC and perhaps linearity too, at least to a small extent.

\subsection{Calibration}
The C-index, as a rank correlation, only evaluates the ability of a model to predict the order of the events. However, a high C-index does not guarantee that the model can also reliably predict the probability of observing the occurrence of an event before a given time. Indeed, arbitrarily rescaling the times (preserving the order) would not change the C-index at all. To evaluate the probability estimation, a model calibration is necessary. The most popular evaluation metric for calibration is the Brier score \cite{Gerds2006}.

\begin{figure}[h!]
    \centering
    \includegraphics[width=1.0\linewidth]{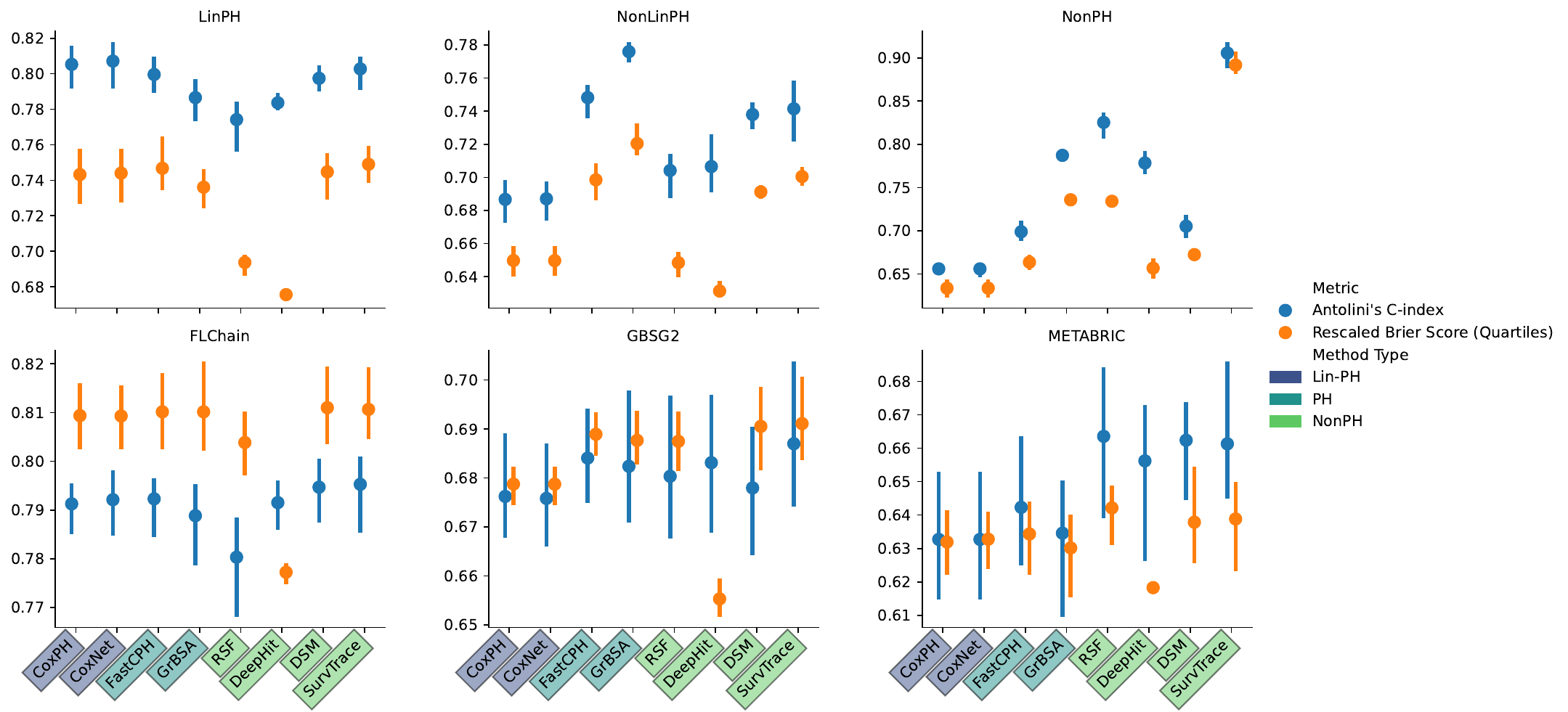}
    \caption{\textbf{Brier score performance performance across all methods and datasets.} Error bars are displayed for the rescaled Brier score (orange bars) and Antolini's C-index (blue bars). Specifically, each dot represents the average score value, while the bars show the range of each score resulting from the 3-fold cross-validation.
    The original Brier score \(BS\) has been rescaled as \(1 - 2BS\) to have the same range and direction of the C-index, allowing an easier visual comparison.
    Methods' assumptions are highlighted with different colors in the horizontal labels.}
    \label{fig:antolini-brier}
\end{figure}

Figure \ref{fig:antolini-brier} displays the Brier scores for all methods across the synthetic and real datasets, compared with the Antolini's C-indices. Full data is available in Supp. Table 1.
Note that we rescaled the Brier score to allow for an easier comparison with the C-index (see figure legend for details).

Generally, we observed that models showing a good Antolini's C-index tend to be more calibrated, with a Pearson Correlation of 0.87 between the two scores. The only systematic exception is DeepHit, that achieved consistently poor Brier scores across all the datasets, ranking last in all but one dataset. Indeed, if we exclude DeepHit, the Pearson correlation between Antolini's C-index and Brier score increases to 0.92. 

In most datasets method ranking is not overly affected by the choice of metrics, including the synthetic datasets, FLChain and GBSG2. However, there are some discrepancies between the two. In Metabric, the Brier score is flatter than the C-index, providing little discrimination between methods. 

\subsection{Time dependent ROC-AUC}
Another commonly used metric is the time-dependent ROC-AUC, which is again based on ranking and is closely related to Antolini's C-index. Indeed, the correlation between the two is very high (0.97) and thus using the ROC-AUC does not provide much additional information with respect to the C-index. Full data is available in Supp. Table 1.

\subsection{Ablation study on the training sample size}

\begin{figure}[h!]
    \centering
    \includegraphics[width=1.0\linewidth]{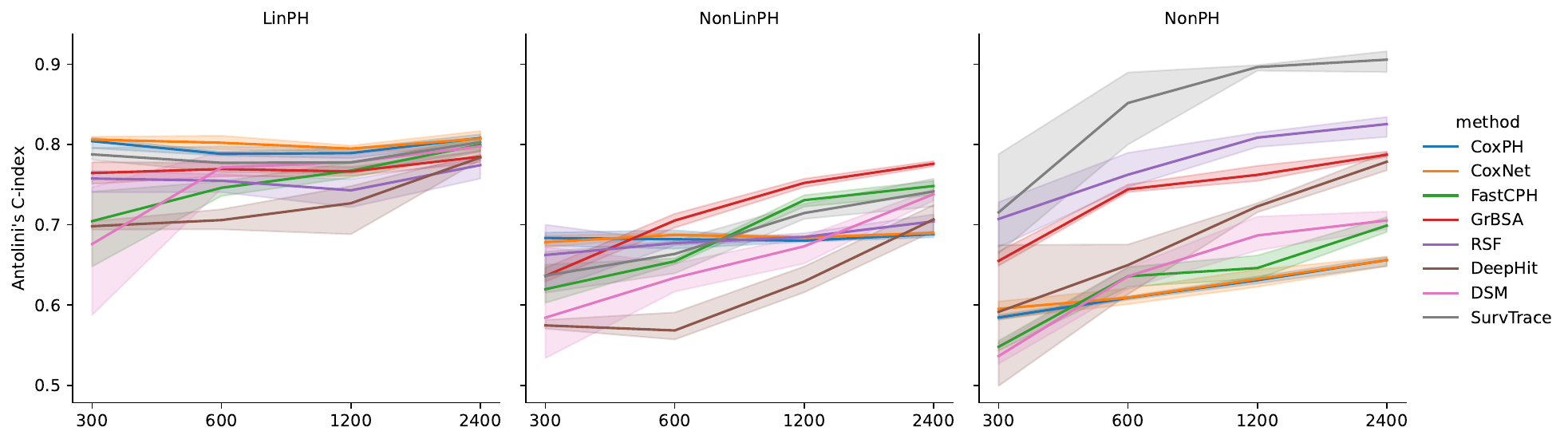}
    \caption{\textbf{Performance of methods at different subsamples of the syntetic datasets.} Y-axis displays the Antolini's C-index, while the x-axis shows the sample size.}
    \label{fig:sample-size}
\end{figure}

In practical applications, sample size has also a crucial role in determining which method will yield the best survival model for a specific dataset. Therefore, we performed an ablation study calculating, for each method, the Antolini's C-index at different subsamples of the synthetic datasets, considering a training sample size ranging between 300 and 2400 samples. In order to factor out the effect of event censoring, for each synthetic data set, the subsamples have been selected from the same pool of 10.000 points in such a way that each subsample contains the same fraction (30\%) of censored events.
Results for the three synthetic datasets are displayed in Figure \ref{fig:sample-size} and reported in Supp. Table 2.

In general, it is reasonable to expect that the performance for all methods will increase at larger sample sizes, since the models are able to learn more complex relations between input features and outcomes. However, saturation effects might occur.  

For the LinPH dataset, the best models are CoxPH and CoxNet and they achieved performance saturation at the lowest sample size of 300, showing that, if their assumptions hold, these models do not require many samples to achieve their best result. Other models like SurvTRACE and GrBSA also showed similar trends, even though at slightly lower C-index values. However, models like FastCPH, DSM and DeepHit performed significantly worse at 300 samples and improved their C-index at higher numbers, but never exceeding the performance obtained by CoxPH and CoxNet.

In the NonLinPH dataset, at 300 samples CoxPH and CoxNet were the best performers, while the other methods, theoretically better suited to the dataset, failed to learn the complex non-linear interactions between features. This changed at higher sample size, where the non-linear models performance continued to grow, overtaking the linear models that already had reached saturation. 
GrBSA only needed 600 samples to surpass CoxPH/CoxNet and become the best performing method, while FastCPH and SurvTRACE overcame the linear models at 1200 samples. DSM, RSF and DeepHit required at least 2400 samples to perform better than the linear approaches. Overall, it seems that the non-linear models did not achieve performance saturation even at the largest sample size of 2400. 

In the NonPH dataset, SurvTRACE outperformed the other methods at all sample sizes. RSF achieved almost the same performance as SurvTRACE at 300 samples, but failed to keep the pace starting from 600 samples, remaining a distant second. Except for SurvTRACE, which appears to achieve saturation at 1200 samples, no other methods showed saturation in this dataset. Interestingly, even both CoxPH and CoxNet methods, whose assumptions are violated here, were able to improve their C-indices with increasing number of samples, differently from the other datasets. It is also notable that these linear models were able to outperform some non-linear and non-PH methods (FastCPH and DSM) at the 300-sample size.

\subsection{Running time}

\begin{figure}
    \centering
    \includegraphics[width=0.7\linewidth]{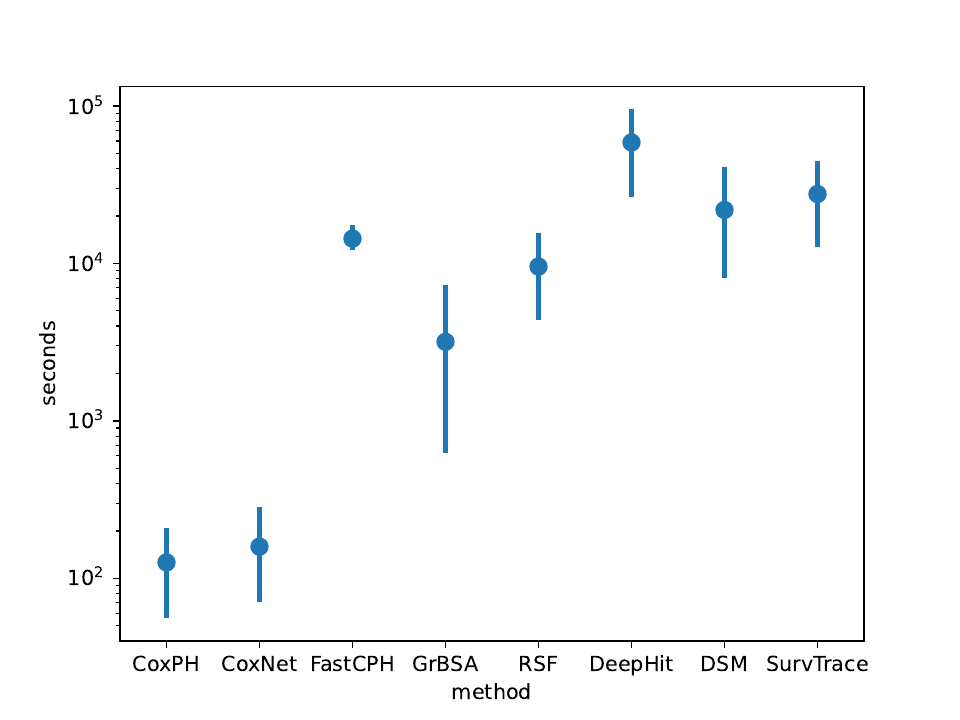}
    \caption{\textbf{Time elapsed for hyper-parameter optimization and training of each method.} The error bars represent the time elapsed in the three real clinical datasets. The vertical axis scale is logarithmic.}
    \label{fig:fitting-time}
\end{figure}

Another factor to consider when applying machine-learning-based methods is the computational cost for training the models, in particular when hyper-parameter optimization and the consequent cross-validation are taken into account. Depending on the available computational resources, a long computational time for training can severely limit the optimization process, reducing the number of hyper-parameter combinations that can be tested. 
Figure \ref{fig:fitting-time} shows the total time required for hyper-parameter optimization and training of each method.
As could be expected, CoxPH and CoxNet, as linear methods, are the fastest methods, requiring only a few minutes. The GrBSA and RSF are the fastest non-linear methods, taking less than one hour and up to three hours on FLChain (the largest real dataset), respectively. The other deep learning methods are even slower, with DeepHit reaching 19 hours on FLChain. For all methods, the larger the dataset, the longer the computational time is, with the exception of FastCPH, which took roughly the same time on all three datasets.

\section{Conclusions}
Linearity and PH assumptions play a critical role in determining the best method for a given survival dataset. When they both hold, there seems to be no reason to go beyond a CoxPH regression, which provides the best performance, while also having the benefits of fast computational training times and a clear, statistically-sound interpretation. In addition, even when linearity and PH assumptions are violated, it can still provide good predictions proving to be remarkably robust, in particular at small sample sizes, where methods allowing for non-linearity and non-PH may not have enough data to converge to a better solution than CoxPH Furthermore, CoxNet shows almost no differences in performance with respect to the CoxPH model, probably due to the high sample-size over feature-number ratio that makes regularization unnecessary even at lower sample sizes. 

When enough samples are available, more complex methods may provide more accurate models when the Cox assumptions are not verified. Since there is no direct way to predict when this occurs, in practical terms we can only try to fit the models and check wether they perform better than Cox. When doing this, it is very important to use appropriate metrics for non-PH methods, such as Antolini's C-index and Brier's score to check the model calibration. Using the original Harrell's metric to assess the C-index might be deleterious, since this metric implicitly assumes a constant risk ranking, a condition not necessarily satisfied by non-PH models. Even though in our study Antolini's C-index and Brier's score often showed concordance in ranking models, we recommend using both, as they evaluate different aspects and occasionally high C-index models happen to be badly calibrated. 

Although no method emerged as the best performer in all datasets, we can make some general considerations. SurvTRACE consistently showed good- or high- performance across all datasets in both Antolini's C-index and Brier's score, even at lower sample sizes, never failing too badly. Considering its performance in the synthetic datasets, this method showed its capability to adapt to rather extreme conditions. 
DSM performed similarly to SurvTRACE in most tests at large dataset sizes, however it performed slightly worse at lower sample sizes. On the other hand, DeepHit never overcame the other methods in Antolini's C-index, performing quite badly at lower sample sizes and in terms of Brier's score, thus providing no advantages with respect to other methods. When considering the whole optimization time, DeepHit also resulted as the slowest method. Finally, non-linear PH methods FastCPH and GrBSA performed similarly, with GrBSA showing better C-indices in some datasets and requiring much less computational resources than FastCPH.

Since the best method in machine learning is always dependent on the dataset, testing different methods is the only way to know which one to use. From our study, we suggest to start testing with CoxNet, GrBSA and SurvTRACE, since other methods do not appear to provide compelling benefits over those three. To address this, we therefore included the benchmark data into the SurvHive package \cite{birolo2025}, providing support for hyperparameter tuning, time-dependent risk evaluation, and cross-validation for censored data.

\section{Funding}
The authors thank the Italian Ministry for Education, University and Research under the programme “Ricerca Locale ex-60\%” and PNRR M4C2 HPC -1.4 “CENTRI NAZIONALI - Spoke 8 for fellowship support. In addition, the authors thank the European Union'’s Horizon 2020 projects Brainteaser (Grant Agreement ID: 101017598)  and GenoMed4All (Grant Agreement ID: 101017549). Dr. Tiziana Sanavia thanks PRIN project "Investigating the role of NF-YA isoform/lncRNA axis in mesoderm specification" of the Italian Ministry for Education, University and Reasearch (Grant ID: 20224TWKNJ).

\bibliography{survhive}

\end{document}